\documentclass[acmtog,nonacm]{acmart}
\usepackage{balance}
\renewcommand\footnotetextcopyrightpermission[1]{} 
\usepackage{booktabs} 
\usepackage{amsmath}
\usepackage{amsthm}
\usepackage{lipsum}
\usepackage{mathtools}
\usepackage{siunitx}

\usepackage{booktabs} 
\usepackage{subcaption}
\usepackage{afterpage}
\usepackage{multirow}
\usepackage{multicol}
\usepackage{overpic}
\usepackage{wrapfig}
\usepackage{graphicx}
\usepackage{soul}

\usepackage{wrapfig}

\def \path {\mathit{path}}

\usepackage{array}

\newcolumntype{L}[1]{>{\raggedright\let\newline\\\arraybackslash\hspace{0pt}}m{#1}}
\newcolumntype{C}[1]{>{\centering\let\newline\\\arraybackslash\hspace{0pt}}m{#1}}
\newcolumntype{R}[1]{>{\raggedleft\let\newline\\\arraybackslash\hspace{0pt}}m{#1}}

\usepackage{pifont}

\newcolumntype{Y}{>{\centering\arraybackslash}X}

\usepackage{xspace}

\usepackage[capitalize]{cleveref}
\crefname{section}{Sec.}{Secs.}
\Crefname{section}{Section}{Sections}
\Crefname{table}{Table}{Tables}
\crefname{table}{Tab.}{Tabs.}

\usepackage[ruled]{algorithm2e} 
\usepackage{algorithmic}
\usepackage[export]{adjustbox}

\SetAlFnt{\small}
\SetAlCapFnt{\small}
\SetAlCapNameFnt{\small}
\SetAlCapHSkip{0pt}
\acmJournal{TOG}

\definecolor{georgecolor}{RGB}{255, 87, 51}

\hypersetup{final}

\begin{document}
\title{PE-Field 4D: Video Generation Models as Canvas}

\author{Yunpeng Bai}
\authornote{This work was conducted during an internship at Pixocial Technology.}
\affiliation{%
  \institution{University of Texas at Austin}
  \country{USA}
}

\author{Haoxiang Li}
\affiliation{%
  \institution{Pixocial Technology}
  \country{USA}
}

\author{Qixing Huang}
\affiliation{%
  \institution{University of Texas at Austin}
  \country{USA}
}

\renewcommand\shortauthors{Bai, Li, and Huang}


\begin{teaserfigure}
    \centering
    \includegraphics[width=\textwidth]{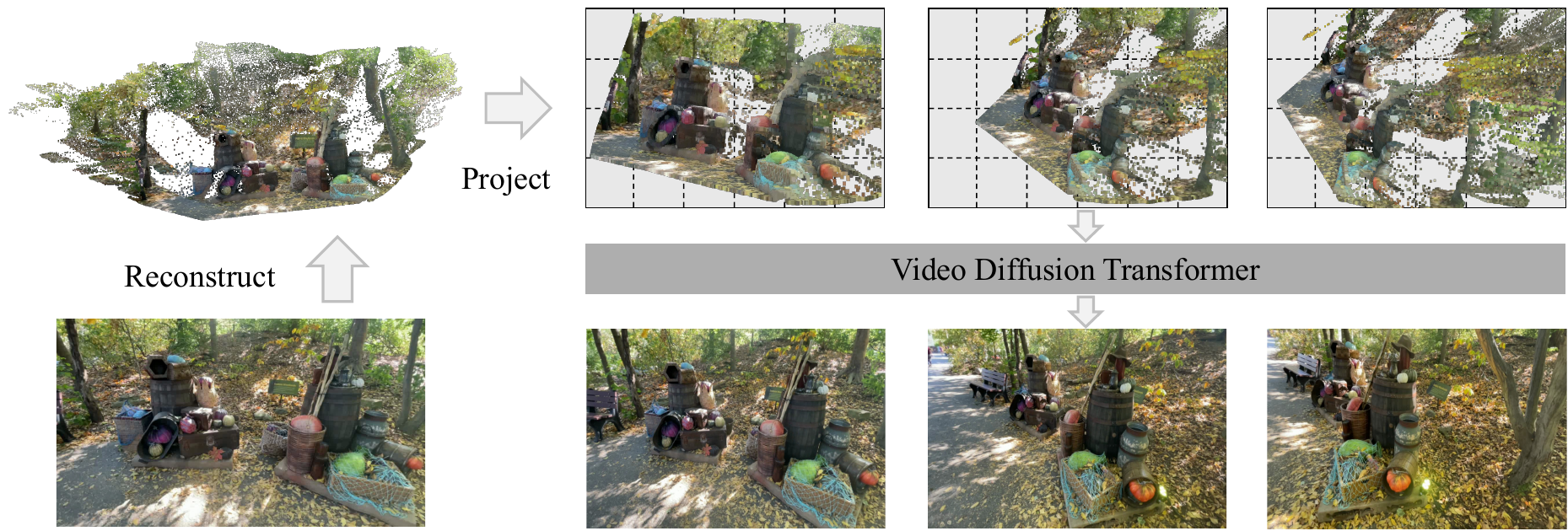}
    \caption{
    Overview of our geometry-aware video generation framework. 
    Given an input image/video, we reconstruct its scene geometry and project reference content onto target camera views to obtain position-aligned context. 
    The video diffusion transformer then uses this structured context to synthesize videos that follow the desired camera trajectory while preserving scene appearance and spatial structure.
    }
    \label{fig:teaser}
\end{teaserfigure}

%
%


%
%




\begin{abstract}
Diffusion Transformers have recently achieved strong performance in video generation, yet controlling scene geometry under viewpoint changes and camera motion remains challenging. In this work, we revisit the role of positional encoding in video diffusion transformers and show that it provides a useful spatial bias for geometry-aware control. Specifically, if reference tokens are encoded according to their projected locations in the target view, the denoising model is encouraged to retrieve content from position aligned regions of the input video. Building on this observation, we introduce a geometry-aware cross-attention mechanism that enables target video latent tokens to attend to structured context tokens derived from reference images or frames. To establish correspondence between the reference content and the target camera trajectory, we equip the context tokens with a projected positional encoding scheme that combines target-view 2D reprojection with depth-aware disambiguation. At the same time, we preserve the original spatiotemporal positional encoding of the generated video latent, allowing geometric guidance to be injected while maintaining consistency with the video model's native latent structure. The resulting framework provides a simple and effective approach for controllable video generation. It improves spatial controllability in viewpoint-dependent editing tasks, including camera re-trajectory, novel-view video synthesis, and geometry-aware video editing, while preserving the generative prior of the underlying video diffusion model. The code is available at: \url{https://github.com/MTLab/PE-Field}.

\end{abstract}

\maketitle

\section{Introduction}
 
Diffusion transformers \cite{peebles2023scalable} have recently emerged as a powerful foundation for visual generation, achieving strong performance in both image and video synthesis. In particular, video diffusion transformers~\cite{zhu2024sora,gao2025seedance,kong2024hunyuanvideo} inherit the expressive modeling capacity of large-scale image generators while extending it to spatiotemporal data, allowing for increasingly realistic video generation from text, images, and other conditions. Despite this progress, controlling the spatial structure of generated videos remains challenging. Existing video generation models can often synthesize visually plausible motion, yet they still struggle to faithfully preserve scene geometry under viewpoint changes, camera motion, or spatial editing. As a result, 4D video editing tasks such as camera re-trajectory, view-consistent generation, and geometry-aware video editing remain difficult for current transformer-based diffusion models. Existing camera re-trajectory methods either use camera poses as generation conditions \cite{bai2025recammaster} or first perform 4D reconstruction and then inpaint the re-rendered results \cite{ren2025gen3c,yu2025trajectorycrafter}. These approaches either lack explicit geometric structure, leading to inaccurate control, or are limited by the quality of the reconstructed scene, and they generally do not generalize well to broader spatial editing tasks.

In this paper, we study 4D generative video editing \cite{lee2025generative} under the video transformers paradigm. Our goal is to leverage rich information in pre-trained models while achieving efficient video editing.  Our approach takes inspiration from recent work on \emph{PE-Field}~\cite{bai2025positional}, which shows that positional encoding can be used as a structured interface for geometry-guided generation. Rather than treating positional encoding as a fixed indexing mechanism, PE-Field demonstrates that modifying the positional codes of source tokens according to their projected locations in a target view, i.e. a \textsl{reference image}, can effectively guide a diffusion transformer to perform novel view synthesis. This result is conceptually significant: it suggests that positional encoding not only organizes token identity but also actively shapes how attention transports information across views. However, PE-Field has only been validated in image generation models, where the target is a single image and the geometry is relatively simple.

Generalizing the PE-field idea to video editing is difficult, as compared with image generation, video generation requires the model to jointly reason about space, time, and motion. In addition, many practical architectures also rely on temporally compressed latent representations. These characteristics introduce new challenges for PE-based geometry control. First, the model must preserve temporal coherence while incorporating spatially structured context from reference frames. Second, the image editing models used by PE-Field naturally contain an attention pathway between the reference branch and the generated latent, whereas mainstream video generation and editing models typically do not provide such a structure that can be directly reused. Third, temporal compression in video VAEs may blur the correspondence between latent tokens and original frames, making precise positional reassignment ambiguous.

Our starting point is a simple reinterpretation of why PE warping works. We argue that the success of PE-Field can be understood as an \emph{attention-guidance mechanism}: when source tokens are assigned positional encodings corresponding to their projected locations in the target view, attention naturally becomes biased toward retrieving position aligned content. From this perspective, PE warping is especially well suited for cross-attention. If a video diffusion transformer is allowed to query reference tokens whose positional codes already reflect target-view geometry, then the model can directly retrieve the source content that should appear at each target token location. This converts positional encoding from a passive indexing scheme into an active geometric routing signal.

Motivated by this view, we propose a geometry-aware extension of PE-Field for video diffusion transformers. To efficiently handle video data, our method augments a standard video diffusion transformer with a cross-attention mechanism that jointly attends to global target latent memory and geometry-aware context memory. The target video latent tokens keep the model's original spatiotemporal positional encoding, preserving the pretrained video prior. In contrast, context tokens extracted from reference images or frames are assigned positional encodings based on their projected target-view coordinates. To further disambiguate tokens that project to similar 2D positions, we introduce a depth-aware positional encoding strategy that injects normalized depth offsets into the temporal axis of the context encoding. In addition, to address ambiguity caused by temporal compression in video latent representations, we modify the context construction process so that each latent frame corresponds to a single source frame, enabling accurate positional reassignment.

Our method provides a simple but effective way to 4D generative video generation based on designing positional encoding fields. By preserving the original denoising dynamics of the video model while equipping context tokens with geometrically meaningful positional encodings, the transformer can use reference frames as structured spatial context rather than merely appearance hints. This leads to improved controllability for tasks such as camera re-trajectory  and view-consistent video synthesis (see Fig.~\ref{fig:teaser}), and more broadly suggests that positional encoding can serve as a general interface for injecting geometric structure into transformer-based generative models.

Our contributions are summarized as follows:
\begin{itemize}
    \item We extend the PE-Field paradigm from image diffusion to video diffusion transformers, showing that warped positional encoding can be used to guide geometry-aware video generation.
    \item We first inject geometric structure directly into the video generation Transformer architecture itself and propose a geometry-aware cross-attention design that combines global target latent memory with grouped local context memory.
    \item We propose a projected positional encoding strategy for context tokens that incorporates target-view 2D reprojection, depth-aware disambiguation, and frame-level correspondence, resolving the temporal compression ambiguity in video latent representations and enabling precise positional reassignment.
    \item Experiments show that our method outperforms existing approaches on camera re-trajectory tasks and further generalizes to broader 4D spatial editing applications.
\end{itemize}
\section{Related Works}

\subsection{Large-Scale Video Synthesis}

Recent progress in video generation has been largely driven by scaling both model capacity and video training data~\cite{zhu2024sora,gao2025seedance,kong2024hunyuanvideo}. 
Diffusion-based video transformers that model a clip as a joint spatiotemporal sequence~\cite{gao2025seedance,kong2024hunyuanvideo,yang2024cogvideox,wan2025wan} have shown impressive synthesis quality and have become a widely adopted design for text-to-video and image-to-video generation. 
However, because the entire sequence must be processed within a single denoising model, memory consumption grows quickly with video length and resolution. 
This makes long-duration generation difficult and often requires long videos to be produced as separate clips, where cross-clip consistency is hard to guarantee. 
Auto-regressive video generators~\cite{huang2025self,chen2024diffusion,zhang2025packing,song2025history,gu2025long} mitigate this issue by generating future segments conditioned on previously generated content. 
Although this strategy extends the reachable temporal span, the model can only access a limited history window, so information from earlier parts of the scene may still be gradually lost.

\subsection{Spatial Memory and Camera Control}

Maintaining a stable scene representation over long time horizons is closely related to camera-controlled generation, since both require the model to synthesize consistent observations of the same scene across changing viewpoints. 
In long-form generation, drift in object appearance, layout, or geometry is commonly observed when the accessible context is limited~\cite{song2025history,decart2024oasis,kanervisto2025world}. 
One family of methods addresses this issue by building an explicit geometric memory from past frames~\cite{wu2025video}. 
These methods estimate 3D structures such as point clouds and integrate them with volumetric fusion techniques such as TSDF~\cite{andy2017tsdf}, which can help preserve coarse scene layout for subsequent generation. 
However, explicit geometry is less convenient for open-ended scenes and may lose detailed appearance or thin structures during reconstruction and fusion.


Camera trajectory control faces a similar alignment problem. 
Existing methods typically either construct explicit guidance from an input image or video, such as point-based renderings~\cite{cao2025uni3c,feng2024i2vcontrol,li2025realcam,ma2025you,you2024nvs,yu2025trajectorycrafter,yu2024viewcrafter,zhai2025stargen}, tracked correspondences~\cite{gu2025diffusion}, or optical flow~\cite{burgert2025go}, or inject camera information into pretrained video diffusion models through additional trainable modules and conditioning representations, including pose parameters~\cite{bai2025recammaster,wang2024motionctrl}, Pl\"{u}cker ray embeddings~\cite{bahmani2025ac3d,he2024cameractrl,he2025cameractrl}, and relative camera encodings~\cite{zhang2025unified}. 
In contrast to these approaches, our work focuses on explicit token-wise spatial alignment between generated tokens and reference observations. 
We use this alignment as a common mechanism for camera control and spatial memory, leading to a unified formulation based on time-aware positional encoding warping.

\subsection{Positional Encoding Designs of DiT}

Several recent papers have studied how to modify RoPE for various 3D vision tasks, examples include CaPE~\cite{li2025cameras} and RayRoPE~\cite{wu2026rayrope}. Closest related to this work is PE-field~\cite{bai2025positional}, which introduces the idea of fine-tuning a pre-trained diffusion transformer based on a novel positional encoding scheme for novel-view synthesis. \cite{gao2026gscompleterdistillationfreepluginmetricaware} applied PE-field to synthesize novel-views for reconstructing 3D Gaussians. Two concurrent papers~\cite{xu2026ucmunifyingcameracontrol,DBLP:journals/corr/abs-2603-17117} extend PE-field to videos. They are mostly suitable for videos of camera motions of static scenes. Our approach differs from them in terms of specific PE strategies, focusing on leveraging information in pre-trained models, efficiency in training and inference, and modeling dynamic scenes.

\section{Method}

\begin{figure}
    \centering
    \includegraphics[width=\linewidth]{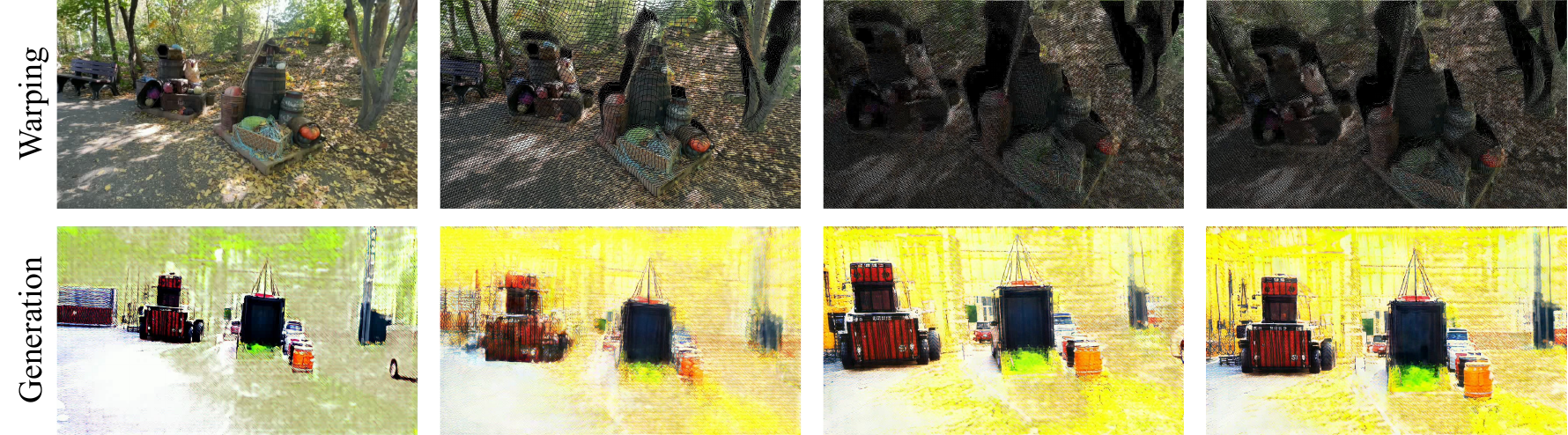}
    \caption{
    Visualization of position-aligned attention in a pretrained video diffusion transformer. 
    We insert cross-attention into the pretrained video model and assign new positional codes to the reference tokens shown in the top row. 
    During generation, these tokens are attended by the video latent tokens according to their assigned positions, leading to the corresponding content appearing at the aligned spatial regions in the generated results shown in the bottom row.
    }
    \label{fig:dit_analysis}
\end{figure}

\subsection{PE-Field Design Principles for Video Diffusion Transformers}

The key insight of PE-Field is that positional encoding does not merely index token identity but also directly shapes how attention routes information. By assigning source tokens with positional codes that correspond to their projected locations in a target view, the model is encouraged to retrieve content from position-aligned regions during denoising. In this work, we extend this paradigm from image diffusion to video diffusion transformers, where generation must simultaneously model spatial structure, temporal coherence, and camera motion. 

To validate this hypothesis in pretrained video DiTs, we inject a cross-attention mechanism into the pretrained model and manually reassign position context tokens. As shown in Fig.~\ref{fig:dit_analysis}, the generated video tends to place the referenced content at the spatial locations specified by these reassigned positional encodings. This provides direct evidence that pretrained video diffusion transformers already exhibit a useful spatial bias: target latent tokens preferentially attend to reference tokens from position-aligned regions. Based on this observation, we augment a standard video diffusion transformer with a geometry-aware cross-attention mechanism. The target video latent tokens preserve the model's original spatiotemporal positional encoding, while context tokens are assigned positional encodings according to their reprojected target-view coordinates and depth-aware ordering. This allows the model to use reference frames as spatially grounded context for camera re-trajectory and spatial aware video generation, while also providing a practical way to address the memory challenge in video modeling.

\begin{figure*}
    \centering
    \includegraphics[width=\linewidth]{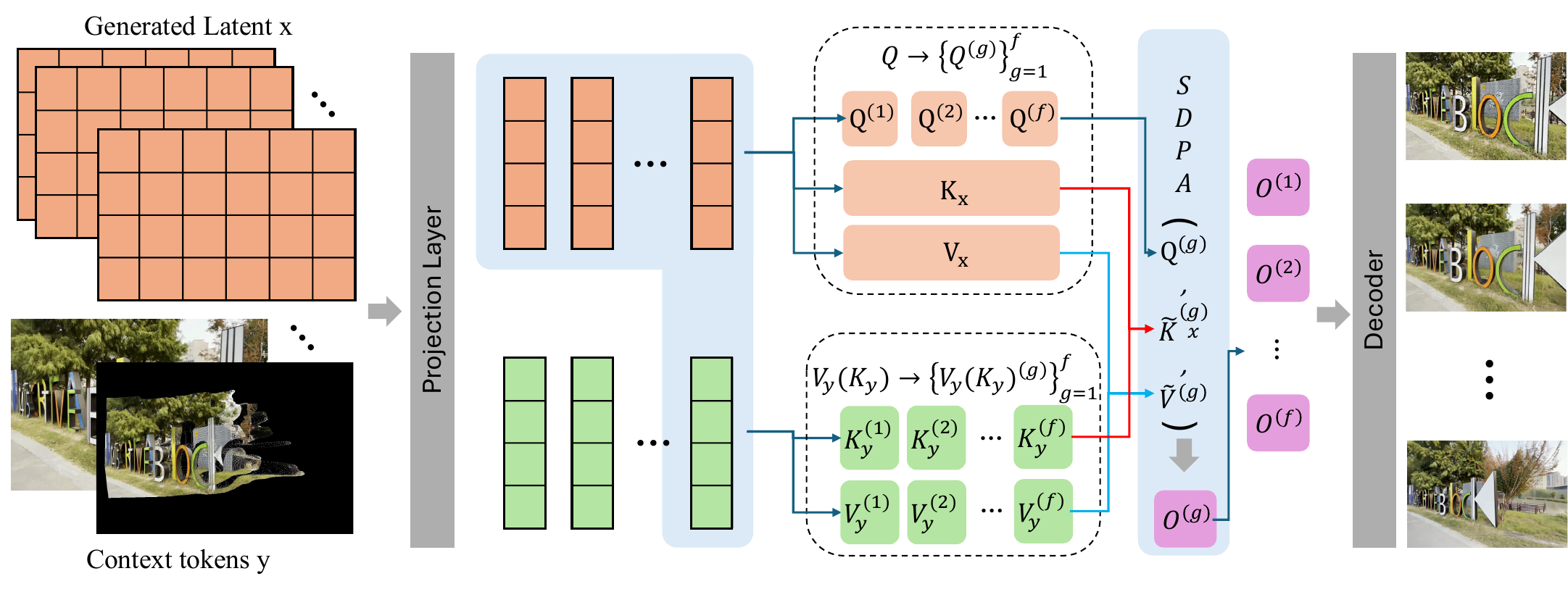}
\caption{
Overview of the proposed geometry-aware cross-attention mechanism for video diffusion transformers.
Given the noisy target video latent tokens \(x\) and the reference context tokens \(y\), we compute queries from the target tokens and construct two key-value branches: a global self-attention memory \((K_x,V_x)\) from the target video latent and a geometry-aware context memory \((K_y,V_y)\) from the reference tokens.
The target queries are divided into groups, while the self-attention memory is shared globally across all groups and the context memory is consumed in a grouped local manner.
For each group \(g\), attention is computed over the concatenated memory \([\tilde{K}^{(g)},\tilde{V}^{(g)}]=[K_x,K_y^{(g)};V_x,V_y^{(g)}]\), allowing each target token to jointly attend to global target-video information and position-aligned reference content.
The context keys are encoded with projected target-view positional encodings, so the reference tokens are retrieved according to where they should appear in the generated view.
}
    \label{fig:architecture}
\end{figure*}

\subsection{Geometry-Aware Cross-Attention}

Let \(x \in \mathbb{R}^{B \times N \times D}\) denote the noisy latent tokens of the target video to be denoised, and let \(y \in \mathbb{R}^{B \times M \times D}\) denote context tokens extracted from reference images or frames. In a standard video diffusion transformer, attention is performed only among the target latent tokens, with positional encoding defined over the three video axes \((t,h,w)\). We preserve this encoding for the target latent tokens and modify only the encoding of the context tokens.

Our attention block augments the original self-attention with an additional geometry-aware context branch. Queries are always computed from the target latent tokens $x$:
\begin{equation}
Q = W_q x,
\end{equation}
while two key-value branches are constructed:
\begin{equation}
K_x = W_k x,\qquad V_x = W_v x,
\end{equation}
\begin{equation}
K_y = W_k y,\qquad V_y = W_v y.
\end{equation}
where $K_x$ and $V_x$ come from latent target video tokens and $K_y$ and $V_y$ come from latent reference frame tokens. After RMS normalization, rotary positional encoding \cite{su2024roformer} is applied to \(Q\) and \(K_x\) using the original video latent PE, while \(K_y\) is encoded using the projected PE to be introduced in Sec.~\ref{sec:projected_pe}. The final memory used by attention is
\begin{equation}
K = [K_x ; K_y], \qquad V = [V_x ; V_y],
\end{equation}
where \([\cdot;\cdot]\) denotes concatenation along the sequence dimension.

This design turns cross-attention into a geometry-aware retrieval process. The target noisy tokens still interact globally through the original self-attention branch, while also querying context tokens whose positional encodings already indicate where they should appear in the target view. As a result, the model can directly select reference content according to the position of the target-view, rather than relying purely on content similarity.

\paragraph{Global self-attention with grouped local context attention.}
A practical issue is that dense attention over all target and context tokens becomes expensive for long video sequences. We therefore use a structured attention pattern in which the self-attention branch remains the same, while the context branch is consumed in a grouped local manner.

Specifically, let \(f\) denote the number of latent frames after temporal compression. 
We divide the sequence into \(f\) groups, each corresponding to one latent frame and containing \(G=N/f\) tokens. We reshape the query tensor into groups,
\begin{equation}
Q \rightarrow \{Q^{(g)}\}_{g=1}^{f}, \qquad Q^{(g)} \in \mathbb{R}^{B \times G \times D},
\end{equation}
The self-attention memory \((K_x,V_x)\) is treated as global and shared across all groups, while the context memory \((K_y,V_y)\) is partitioned group-wise:
\begin{equation}
K_y \rightarrow \{K_y^{(g)}\}_{g=1}^{f}, \qquad
V_y \rightarrow \{V_y^{(g)}\}_{g=1}^{f}.
\end{equation}
For each group \(g\), attention is computed against the concatenated memory
\begin{equation}
\tilde{K}^{(g)} = [K_x ; K_y^{(g)}], \qquad
\tilde{V}^{(g)} = [V_x ; V_y^{(g)}].
\end{equation}
The group output is then
\begin{equation}
O^{(g)} = \mathrm{SDPA}\!\left(Q^{(g)}, \tilde{K}^{(g)}, \tilde{V}^{(g)}\right),
\end{equation}
where \(\mathrm{SDPA}(\cdot)\) denotes scaled dot-product attention. In this way, the cross-attention to context tokens is restricted to the content corresponding to the same latent frame. The outputs of all groups are finally merged back into the original token order.


This hybrid formulation preserves long-range global reasoning through the target latent tokens, while keeping the geometry conditioned context lookup more spatially localized and efficient. Importantly, this is not a separate module from cross-attention, but the cross-attention mechanism itself: each target query attends jointly to global target latent memory and grouped geometry-aware context memory within a single attention operator.

\subsection{Projected Positional Encoding for Context Tokens}
\label{sec:projected_pe}

Another main design lies in how context tokens are positionally encoded. In a standard video diffusion transformer, positional encoding is defined over the latent video grid and factorized along time and space. We leave this encoding unchanged for the target latent tokens \(x\). However, for context tokens \(y\), we replace their original source-view coordinates with projected target-view coordinates.

For each context token, we compute its 2D projection onto the target camera plane using the source-view geometry. Denote the projected position by $(\tilde{h}, \tilde{w})$.
Instead of assigning the token a positional code based on its original source-frame location, we encode it using this projected target-plane location. Thus, context tokens are indexed according to where they should appear in the generated target view. This follows the central PE-Field principle, now adapted to video generation. By changing only the positional code while leaving the token content unchanged, we bias attention to associate target noisy tokens with context tokens that are position aligned in the target view.

\paragraph{Depth-aware positional disambiguation.}
Projected 2D coordinates alone are not always sufficient, since multiple tokens may land near the same target-plane position while lying at different depths. To distinguish such cases, we additionally encode depth into the positional code.
Rather than introducing a separate depth axis, we incorporate depth into the same positional encoding framework by adding a depth-dependent offset to the temporal axis of the context token. Let \(t\) denote the frame index of the context token and \(d\) its depth. We define an effective temporal coordinate
\begin{equation}
t' = t + \Delta(d),
\end{equation}
where \(\Delta(d)\) is a normalized depth offset. In practice, we normalize depth to the interval \([0, 0.1]\). As a result, tokens originating from frame \(t\) occupy a temporal encoding range within \([t, t+\frac{1}{10}]\), which preserves frame-level separation while introducing an additional front-to-back ordering among tokens within the same frame.

The final positional encoding of a context token is, therefore, constructed from
\begin{equation}
(t + \Delta(d), \tilde{h}, \tilde{w}).
\end{equation}
This design allows the model to distinguish tokens that are close in projected 2D position but belong to different depth layers, while still keeping the overall temporal structure compatible with the original video transformer.

\subsection{Resolving Temporal Compression Ambiguity in Context Latents}

A practical complication arises from the latent representation used in video diffusion models. Many video VAEs temporally compress the input sequence, for example by merging several consecutive RGB frames into a single latent frame. In such cases, the exact contribution of each original frame to a compressed latent token is unknown. This poses a problem for our method, since a single compressed latent frame would correspond to multiple projected positional encodings from different original frames, making the positional assignment inherently ambiguous.

To resolve this ambiguity, we modify the context construction procedure. Instead of feeding distinct consecutive frames directly into the video encoder, we repeat each source frame multiple times before compression, such that each resulting latent frame corresponds to only one original frame.  After encoding, each latent frame therefore contains information from a single original frame rather than a mixture of multiple frames.


This design also requires aligning the temporal coordinates of context and generated latents during cross-attention. 
While each generated latent corresponds to a compressed temporal interval, e.g., four RGB frames, our repeated context latents correspond to individual source frames within that interval. 
We therefore normalize the context-frame indices to the same latent-time range as the generated tokens. 
For example, the four source frames associated with one generated latent step are assigned fractional temporal coordinates \(1, 1.25, 1.5,\) and \(1.75\). 
This fractional indexing preserves the temporal order of individual source frames while ensuring that their positional encodings remain compatible with the compressed latent timeline of the video model.


\section{Experimental Results}

\begin{figure*}
    \centering
    \includegraphics[width=\linewidth]{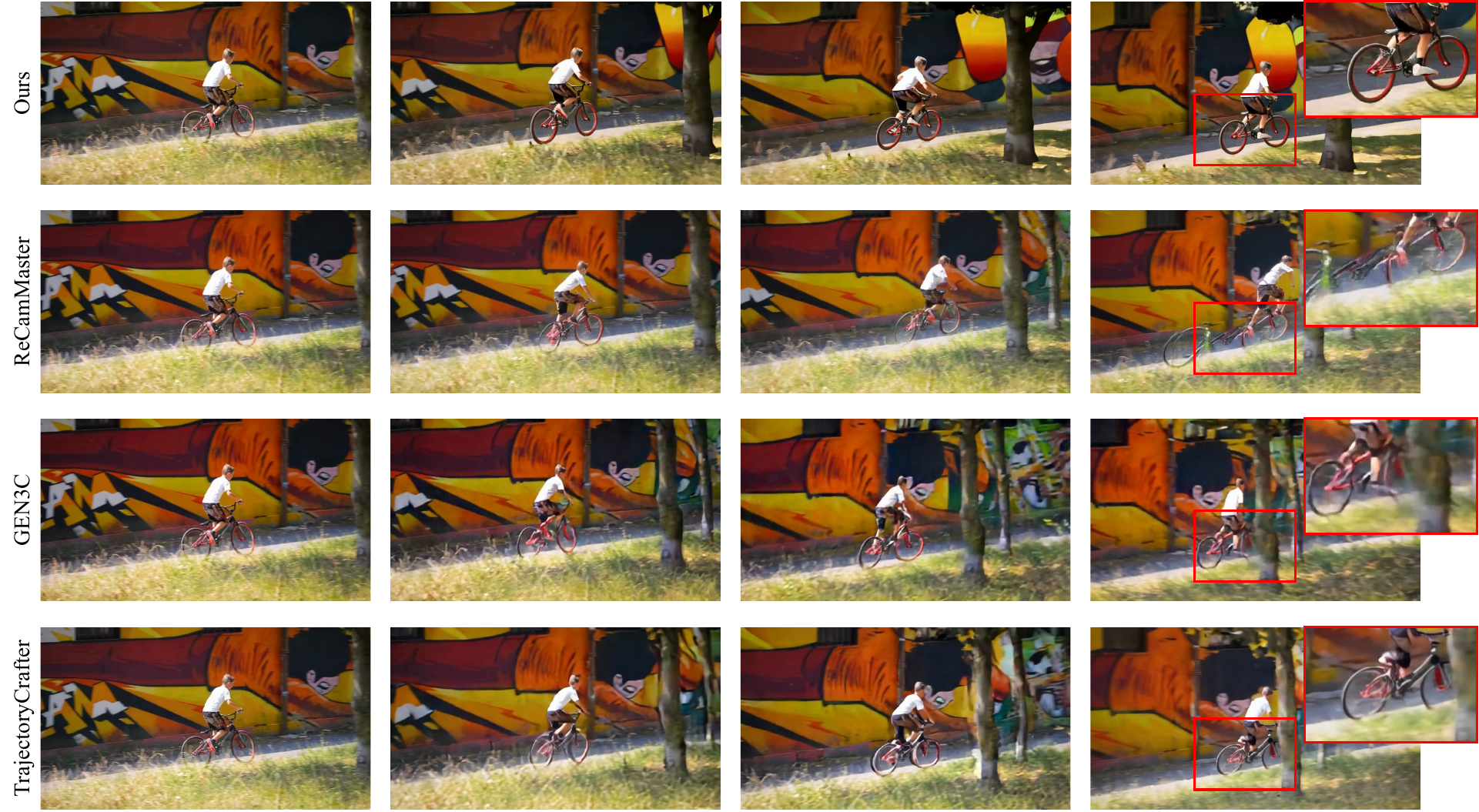}
\caption{
Qualitative comparison on the \textit{bmx-trees} video from DAVIS \cite{pont20172017}. 
Under the transformed camera trajectory, our method preserves more accurate scene geometry and fine structures, such as the bicycle frame, wheels, and thin foreground/background elements. 
Compared with TrajectoryCrafter, ReCamMaster, and GEN3C, our results show better geometric consistency and fewer distortions in detailed regions during viewpoint changes.
}
\label{fig:com_1}
\end{figure*}

\begin{table*}
\centering
\small
\setlength{\tabcolsep}{5.2pt}
\renewcommand{\arraystretch}{1.12}
\resizebox{\textwidth}{!}{%
\begin{tabular}{lcccccccccc}
\toprule
\multirow{2}{*}{Method} 
& \multicolumn{6}{c}{Visual Quality$\uparrow$}
& \multicolumn{2}{c}{Geometric Consistency}
& \multicolumn{2}{c}{Camera Accuracy} \\
\cmidrule(lr){2-7}
\cmidrule(lr){8-9}
\cmidrule(lr){10-11}
& Subject 
& Background 
& Aesthetic 
& Imaging 
& Temporal 
& Motion 
& Dyn-MEt3R$\uparrow$ 
& MEt3R$\downarrow$ 
& TransErr$\downarrow$ 
& RotErr$\downarrow$ \\
& Consistency
& Consistency
& Quality
& Quality
& Flickering
& Smoothness
& 
& 
& 
& \\
\midrule
ReCamMaster 
& 0.8983 & 0.9182 & 0.5052 & 0.6448 & 0.9543 & 0.9752 
& 0.7721 & 0.3585 & 0.0301 & 2.412 \\

GEN3C 
& 0.9011 & 0.9202 & 0.5097 & 0.6259 & 0.9421 & 0.9673
& 0.7426 & 0.3462 & 0.0548 & 6.728 \\

TrajectoryCrafter 
& 0.8824 & 0.9146 & 0.5031 & 0.6013 & 0.9319 & 0.9605
& 0.7315 & 0.3328 & 0.0684 & 8.943 \\

ReDirector 
& 0.8977 & 0.9164 & 0.5078 & 0.6521 & 0.9527 & 0.9784
& 0.8041 & 0.3186 & 0.0189 & 2.107 \\
\midrule

w/o Temporal Dis.
& 0.8946 & 0.9138 & 0.5061 & 0.6417 & 0.9435 & 0.9688
& 0.7564 & 0.3479 & 0.0317 & 3.284 \\

w/o Depth
& 0.9069 & 0.9228 & 0.5171 & 0.6675 & 0.9661 & 0.9864
& 0.8126 & 0.3047 & 0.0161 & 1.982 \\
Ours 
& \textbf{0.9098} & \textbf{0.9247} & \textbf{0.5196} & \textbf{0.6731} 
& \textbf{0.9704} & \textbf{0.9892}
& \textbf{0.8235} & \textbf{0.2968} & \textbf{0.0142} & \textbf{1.887} \\
\bottomrule
\end{tabular}
}
\caption{
Quantitative evaluation on DAVIS videos using camera trajectories from ReCamMaster.
We compare against representative camera-controlled video generation and retargeting methods across visual quality, geometric consistency, and camera accuracy metrics. 
Our method achieves the best overall performance, showing stronger preservation of appearance, more stable geometry under viewpoint changes, and more accurate adherence to the target camera motion.
}
\label{tab:davis_quantitative}
\end{table*}

\subsection{Implementation Details}

We perform our experiments mainly on the 14B versions of the Wan2.1 T2V models \cite{wan2025wan}. To adapt the model to the newly inserted cross-attention modules, we perform LoRA fine-tuning. Specifically, LoRA adapters are inserted into the major projection and feed-forward layers, including the \(q\), \(k\), \(v\), \(o\), and \(\mathrm{ffn}\) layers. We set the LoRA rank to 32 for all experiments. The model is trained with a learning rate of \(1\times10^{-4}\).

Our training data are primarily constructed from the MultiCam dataset \cite{bai2025recammaster}, which provides multi-view video sequences suitable for learning viewpoint-dependent generation and camera re-trajectory control. Since the data set does not provide ground-truth depth maps, we estimate the depth and corresponding camera parameters using ViPE \cite{huang2025vipe}, and use them as input of geometric conditions to construct the projected positional encodings and geometry-aware context tokens.

\subsection{Generation Results} 

Following the evaluation protocol of recent camera-controlled video generation methods \cite{park2025redirector}, we build a test benchmark from 90 videos in the DAVIS dataset and use target camera trajectories provided by ReCamMaster~\cite{bai2025recammaster}. 
The selected videos contain diverse object motions, scene layouts, and fine geometric structures, allowing us to evaluate both visual fidelity and geometry-aware camera control under challenging trajectory transformations. 
For each input video, we generate novel-view results conditioned on the target camera motion and compare our method with ReCamMaster~\cite{bai2025recammaster}, GEN3C~\cite{ren2025gen3c}, TrajectoryCrafter~\cite{yu2025trajectorycrafter}, and ReDirector~\cite{park2025redirector}.

Following the evaluation setting of ReDirector~\cite{park2025redirector}, we assess the generated videos from three complementary aspects: visual quality, geometric consistency, and camera-control accuracy. 
Specifically, we adopt VBench \cite{vbench} metrics to evaluate perceptual and temporal quality, use Dyn-MEt3R \cite{park2025steerx} to measure the geometric consistency of the generated retargeted videos, and compute per-frame MEt3R \cite{asim2025met3r} to evaluate consistency between the generated results and the input video. In addition, we report translation and rotation errors, denoted as TransErr and RotErr, which quantify how well the generated video follows the target camera motion based on relative pose estimation from ViPE \cite{huang2025vipe}.

As shown in Tab.~\ref{tab:davis_quantitative}, our method consistently outperforms previous approaches in visual quality, geometric consistency, and camera precision. 
Compared with ReCamMaster and TrajectoryCrafter, our approach better preserves subject and background appearance while reducing geometric distortions during large camera changes. 
Compared to GEN3C, our method produces a more stable scene structure and more accurate camera follow-up. 
Although ReDirector provides strong camera control, our method further improves both Dyn-MEt3R and MEt3R, indicating better geometry preservation in the generated retakes. 
The lower translation and rotation errors also demonstrate that our results follow the target trajectories more faithfully.

\subsection{Applications}

Beyond camera trajectory transformation, our framework naturally extends to other spatially aware generation tasks, such as animating a single image into a video. 
The key reason is that our method does not rely on task-specific supervision but instead injects spatially aligned context through projected positional encodings. 
When the input is a single reference image, its visual tokens can still be treated as geometry-aware context, allowing the denoising model to retrieve content from spatially corresponding regions while generating temporally coherent motion.

\begin{figure}
    \centering
    \includegraphics[width=\linewidth]{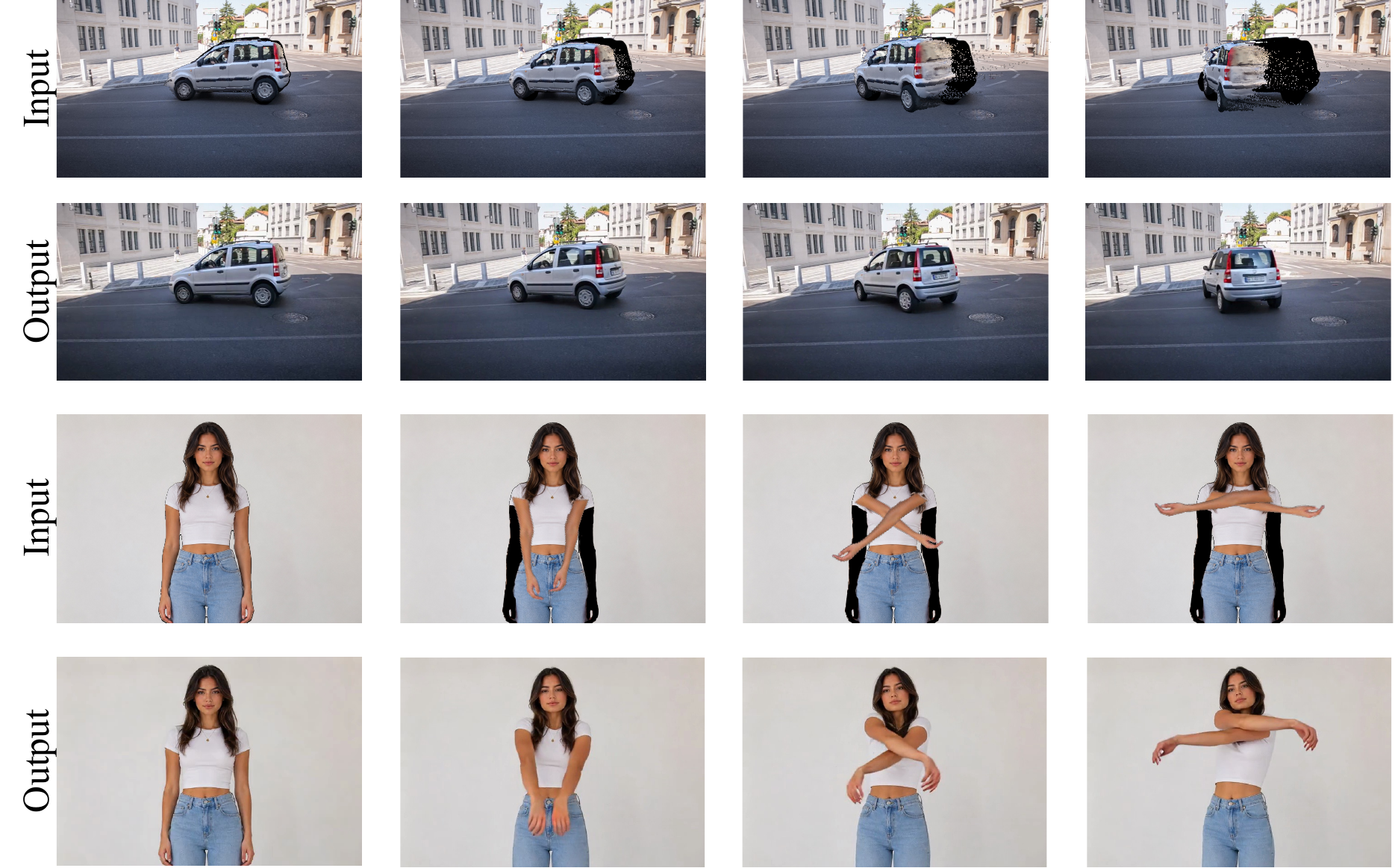}
\caption{
\textbf{General applications beyond camera trajectory transformation.} Our method not only supports controllable camera trajectory transformation, but also naturally extends to other spatially aware generation tasks, such as object motion synthesis and pose-guided animation, by aligning the generated content with structured geometric guidance.
}
\label{fig:image_to_video}
\end{figure}

As shown in Fig.~\ref{fig:image_to_video}, our method can animate static images while preserving the spatial layout and fine structures of the reference content. 
Compared with generation driven only by global image conditioning, the proposed position-aligned context provides more explicit spatial guidance, helping the model maintain object identity, local appearance, and scene structure during motion synthesis. 
This demonstrates that our approach is not limited to camera-controlled video retargeting but can serve as a general mechanism for spatially grounded video generation.

\begin{figure}
    \centering
    \includegraphics[width=\linewidth]{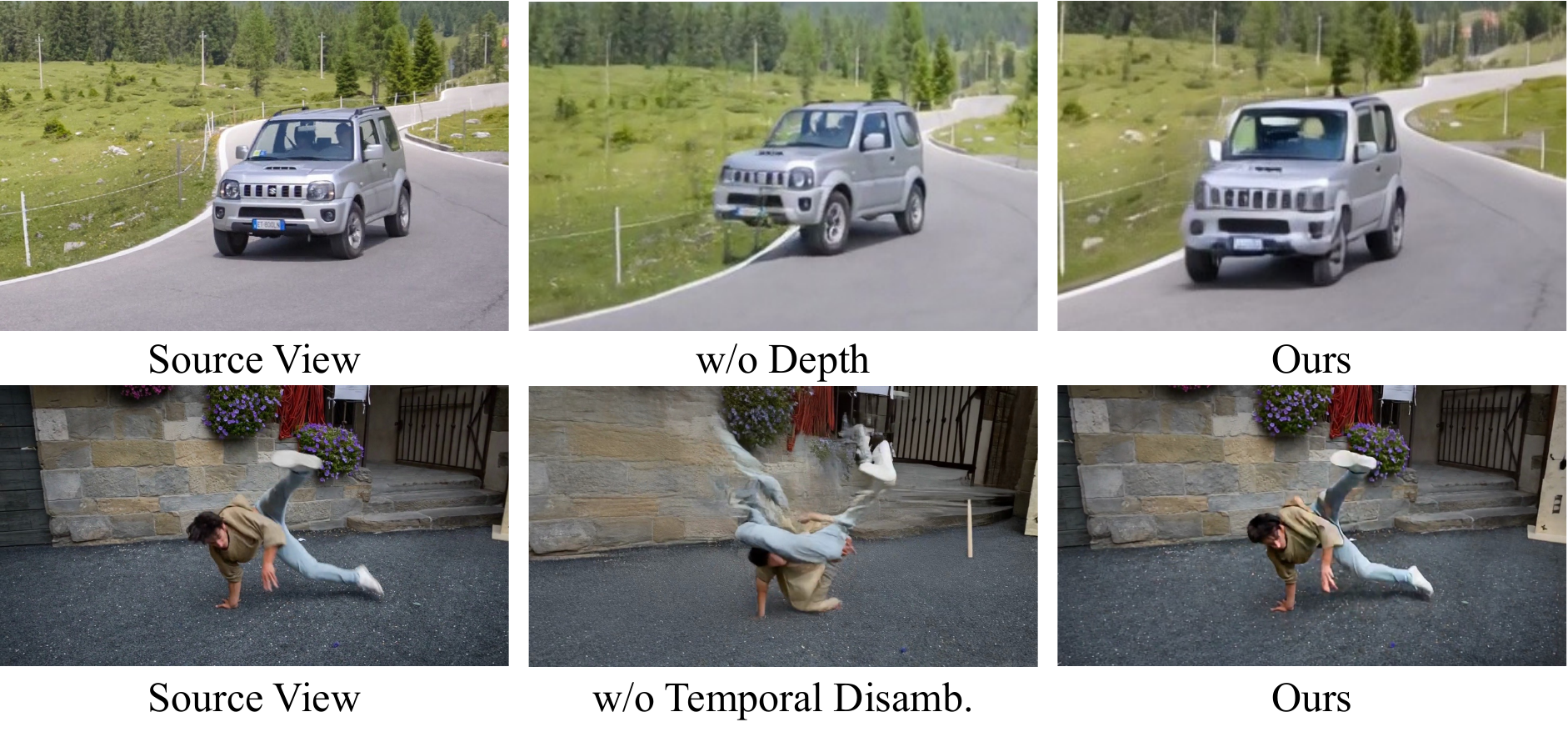}
\caption{
Ablation study on depth and temporal disambiguation. 
Compared with variants without depth-aware or temporal disambiguation, our full method preserves more accurate geometry and achieves more faithful generation under camera transformation.
}
\label{fig:ablation_temporal_compression}
\end{figure}

\subsection{Ablation Studies}

\paragraph{Depth-aware positional disambiguation.}
We ablate the depth-aware offset in our positional encoding. 
Without depth, tokens are distinguished only by projected 2D coordinates, which becomes ambiguous when foreground and background regions overlap in the target view. 
As shown in Fig.~\ref{fig:ablation_temporal_compression}, this leads to less accurate geometry and more distortions near fine structures and occlusion boundaries. 
The quantitative results in Tab.~\ref{tab:davis_quantitative} also show a consistent drop after removing depth-aware disambiguation.

\paragraph{Temporal compression ambiguity.}
We ablate our frame-repetition strategy by directly using the original video VAE compression, where several RGB frames are encoded into one context latent frame. 
This mixes information from multiple source frames and makes projected positional assignment ambiguous. 
As shown in Fig.~\ref{fig:ablation_temporal_compression}, this baseline fails to localize reference content accurately, resulting in weaker geometric alignment. 
Tab.~\ref{tab:davis_quantitative} further shows a clear drop in geometric consistency and camera accuracy, confirming the importance of precise frame-level positional assignment.

\section{Conclusions}

We presented an extension of the PE-Field paradigm from image diffusion to video diffusion transformers. 
Our key idea is to interpret warped positional encoding as a geometry-aware attention signal, enabling target video latent tokens to retrieve structured information from reference images or frames through cross-attention based on aligned position. Our results indicate that positional encoding provides a useful and lightweight interface for incorporating geometric cues into video diffusion transformers.


\paragraph{Limitations.}
Our method introduces additional context tokens, especially when each source frame is encoded separately to resolve temporal compression ambiguity. 
This increases both computational cost and memory consumption during cross-attention, which may limit scalability to longer videos or higher-resolution settings. 
In addition, our current design relies on estimated geometry, such as depth and camera poses, so errors in these estimates may affect the quality of the projected positional encoding and lead to imperfect spatial alignment.

\paragraph{Future work.}
To improve efficiency, one possible direction is to explore more compact context representations instead of directly using temporally expanded video latents as context tokens. 
For example, future work could design lightweight geometry-aware context features that preserve the necessary spatial and appearance information while reducing token count and attention cost. 


\bibliographystyle{ACM-Reference-Format}
\bibliography{ref2}

\newpage 
\appendix


\begin{figure*}
    \centering
    \includegraphics[width=\linewidth]{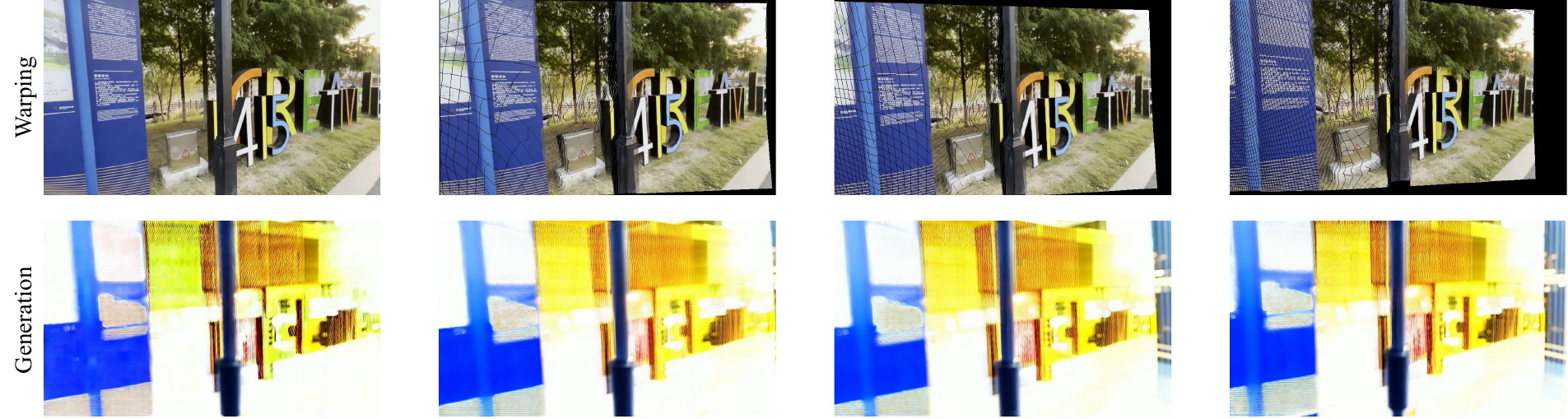}
\caption{
Additional visualization of position-aligned attention in a pretrained video diffusion transformer.
After assigning projected positional codes to context tokens, each target latent attends to the context content according to its projected location in the target view.
}
    \label{fig:supp1}
\end{figure*}

\begin{figure*}
    \centering
    \includegraphics[width=\linewidth]{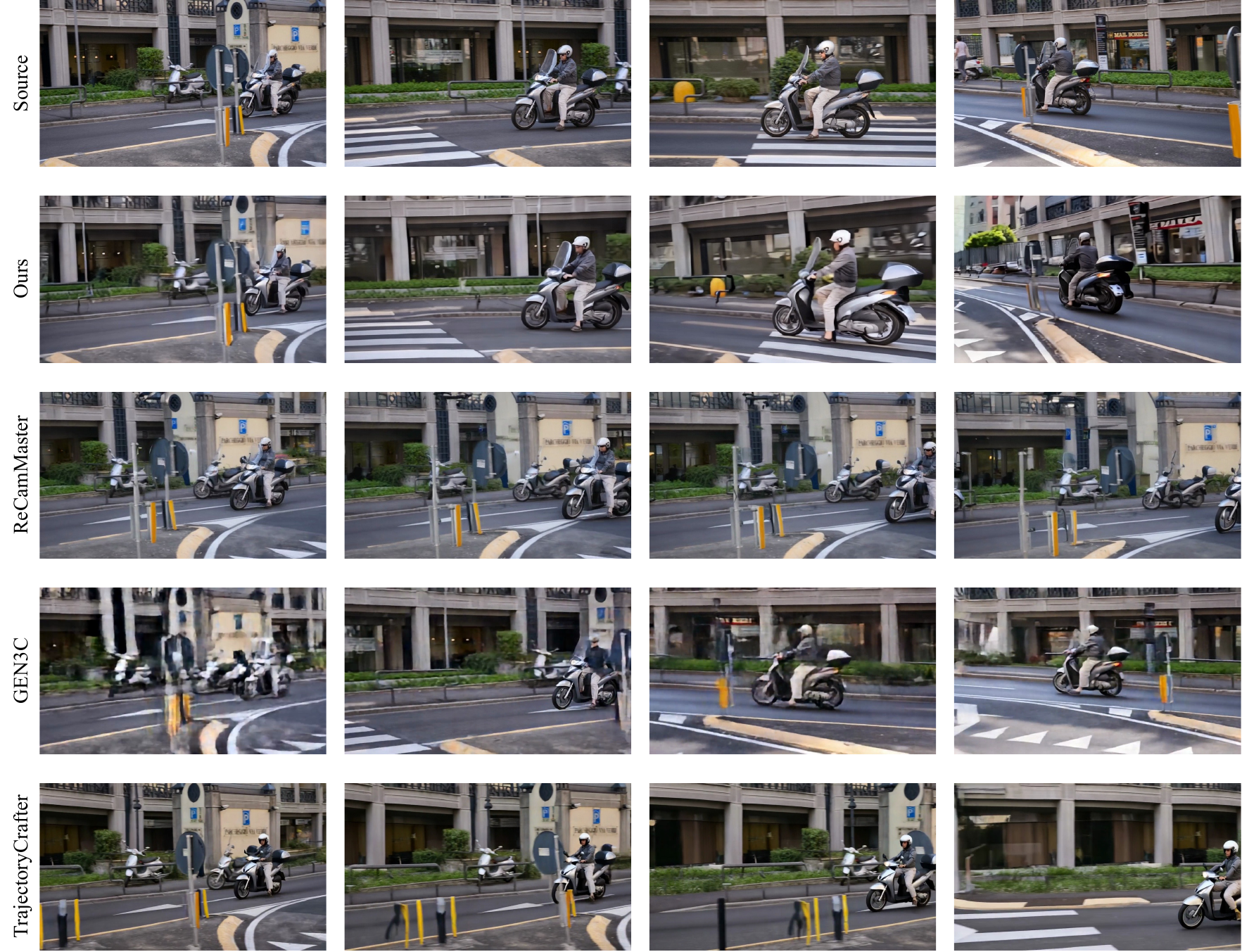}
\caption{
Additional qualitative comparisons. Our method better preserves object structure under large camera changes and produces more position-aligned results than prior methods.
}
    \label{fig:supp2}
\end{figure*}

\begin{figure*}
    \centering
    \includegraphics[width=0.9\linewidth]{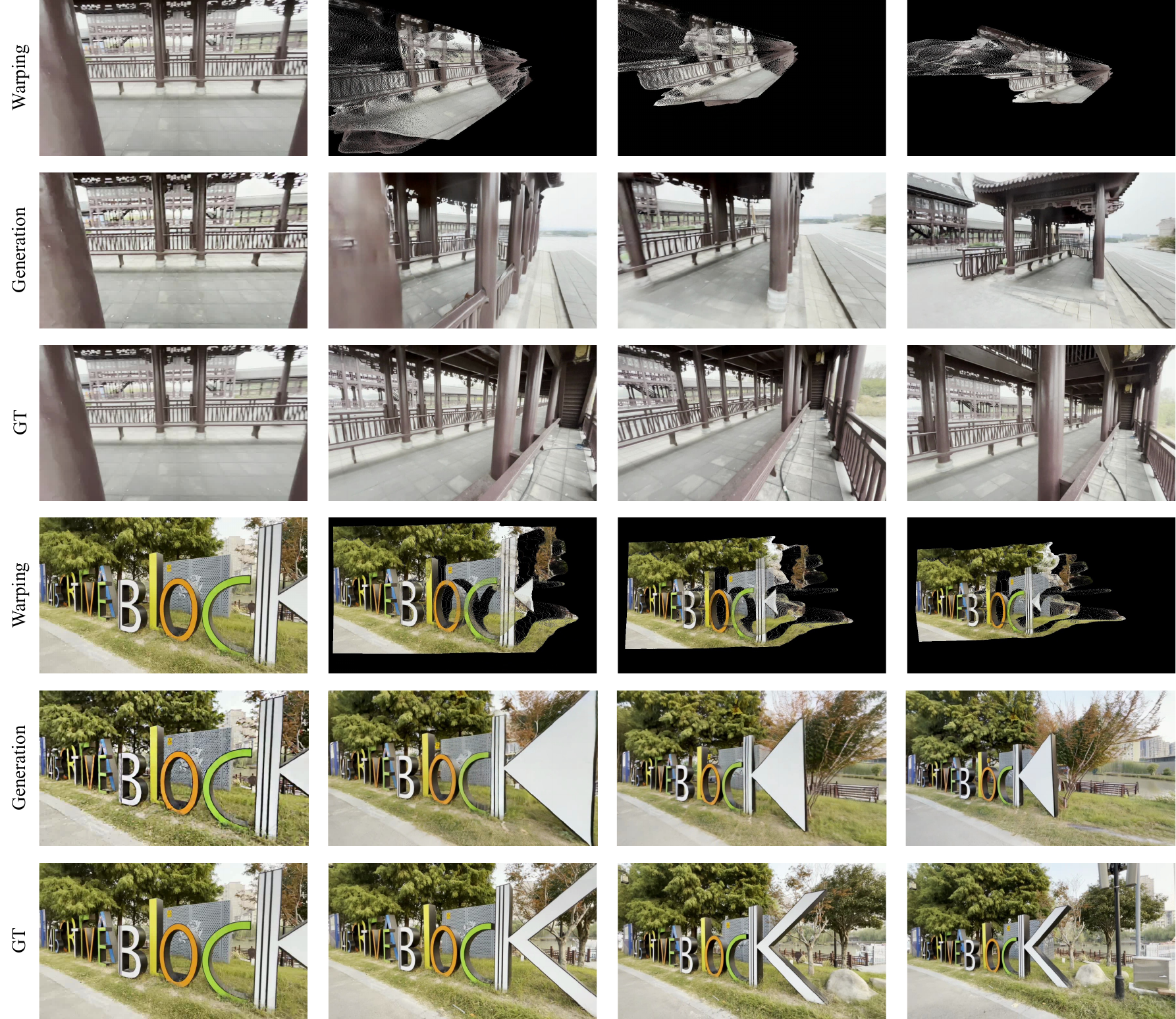}
\caption{
Image-to-video generation results of our method on the DL3DV dataset. The leftmost image is the input image.
}
    \label{fig:supp3}
\end{figure*}

\begin{figure*}
    \centering
    \includegraphics[width=0.9\linewidth]{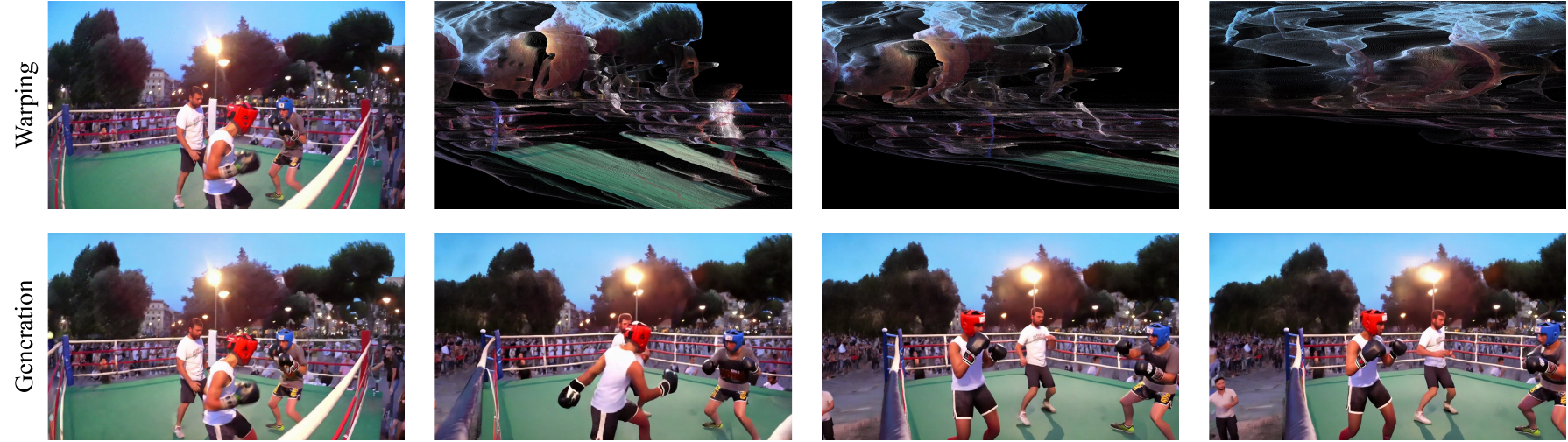}
\caption{
This example shows a case where the reconstructed geometry is inaccurate. Our method can still generate plausible video results even when the reconstruction is noisy or erroneous.
}
    \label{fig:supp4}
\end{figure*}

\end{document}